\documentclass[11pt]{article}
\usepackage[a4paper]{geometry}

\usepackage{amsmath, amsthm, amssymb}
\usepackage{graphicx}
\usepackage{subcaption}
\usepackage{float}
\usepackage[round]{natbib}
\usepackage{url}
\usepackage[hidelinks,colorlinks=true,linkcolor=blue,citecolor=blue,urlcolor=blue]{hyperref} 
\usepackage{booktabs} 
\usepackage{setspace}

\newcommand{\rdsurvival}{\texttt{rdsurvival~}}

\newcommand{\taurisk}{\pi^h}
\newcommand{\taurms}{\tau^h}

\newcommand{\EE}[2][]{\mathbb{E}_{#1}\left[#2\right]}
\newcommand{\PP}[2][]{\mathbb{P}_{#1}\left[#2\right]}
\newcommand\indep{\protect\mathpalette{\protect\independenT}{\perp}}
\def\independenT#1#2{\mathrel{\rlap{$#1#2$}\mkern2mu{#1#2}}}
\newcommand{\cond}{\,\big|\,}

\theoremstyle{plain}

\theoremstyle{definition}

\newtheorem{defi}{Definition}
\newtheorem{assu}{Assumption}
\newtheorem{rema}{Remark}
\newtheorem{alg}{Algorithm}

\makeatletter
\newcommand*\if@single[3]{%
  \setbox0\hbox{${\mathaccent"0362{#1}}^H$}%
  \setbox2\hbox{${\mathaccent"0362{\kern0pt#1}}^H$}%
  \ifdim\ht0=\ht2 #3\else #2\fi
  }
\newcommand*\rel@kern[1]{\kern#1\dimexpr\macc@kerna}
\newcommand*\widebar[1]{\@ifnextchar^{{\wide@bar{#1}{0}}}{\wide@bar{#1}{1}}}
\newcommand*\wide@bar[2]{\if@single{#1}{\wide@bar@{#1}{#2}{1}}{\wide@bar@{#1}{#2}{2}}}
\newcommand*\wide@bar@[3]{%
  \begingroup
  \def\mathaccent##1##2{%
    \if#32 \let\macc@nucleus\first@char \fi
    \setbox\z@\hbox{$\macc@style{\macc@nucleus}_{}$}%
    \setbox\tw@\hbox{$\macc@style{\macc@nucleus}{}_{}$}%
    \dimen@\wd\tw@
    \advance\dimen@-\wd\z@
    \divide\dimen@ 3
    \@tempdima\wd\tw@
    \advance\@tempdima-\scriptspace
    \divide\@tempdima 10
    \advance\dimen@-\@tempdima
    \ifdim\dimen@>\z@ \dimen@0pt\fi
    \rel@kern{0.6}\kern-\dimen@
    \if#31
      \overline{\rel@kern{-0.6}\kern\dimen@\macc@nucleus\rel@kern{0.4}\kern\dimen@}%
      \advance\dimen@0.4\dimexpr\macc@kerna
      \let\final@kern#2%
      \ifdim\dimen@<\z@ \let\final@kern1\fi
      \if\final@kern1 \kern-\dimen@\fi
    \else
      \overline{\rel@kern{-0.6}\kern\dimen@#1}%
    \fi
  }%
  \macc@depth\@ne
  \let\math@bgroup\@empty \let\math@egroup\macc@set@skewchar
  \mathsurround\z@ \frozen@everymath{\mathgroup\macc@group\relax}%
  \macc@set@skewchar\relax
  \let\mathaccentV\macc@nested@a
  \if#31
    \macc@nested@a\relax111{#1}%
  \else
    \def\gobble@till@marker##1\endmarker{}%
    \futurelet\first@char\gobble@till@marker#1\endmarker
    \ifcat\noexpand\first@char A\else
      \def\first@char{}%
    \fi
    \macc@nested@a\relax111{\first@char}%
  \fi
  \endgroup
}
\makeatother

\author{Maximilian Schuessler \\ \texttt{maxsc@stanford.edu}
\and
Erik Sverdrup\\ \texttt{erik.sverdrup@monash.edu}
\and
Robert Tibshirani \\ \texttt{tibs@stanford.edu}
\and
Stefan Wager \\ \texttt{swager@stanford.edu}
}

\title{Nonparametric Regression Discontinuity Designs\\ with Survival Outcomes}

\begin{document}

\maketitle

\begin{abstract}
Quasi-experimental evaluations are central for generating real-world causal evidence and complementing insights from randomized trials. The regression discontinuity design (RDD) is a quasi-experimental design that can be used to estimate the causal effect of treatments that are assigned based on a running variable crossing a threshold. Such threshold-based rules are ubiquitous in healthcare, where predictive and prognostic biomarkers frequently guide treatment decisions. However, standard RD estimators rely on complete outcome data, an assumption often violated in time-to-event analyses where censoring arises from loss to follow-up. To address this issue, we propose a nonparametric approach that leverages doubly robust censoring corrections and can be paired with existing RD estimators. Our approach can handle multiple survival endpoints, long follow-up times, and covariate-dependent variation in survival and censoring. We discuss the relevance of our approach across multiple areas of applications and demonstrate its usefulness through simulations and the prostate component of the Prostate, Lung, Colorectal and Ovarian (PLCO) Cancer Screening Trial where our new approach offers several advantages, including higher efficiency and robustness to misspecification. We have also developed an open-source software package, \texttt{rdsurvival}, for the \texttt{R} language.
\end{abstract}

\section{Introduction}
The regression discontinuity design (RDD) is a quasi-experimental design used to estimate causal effects from observational data. Specifically, the method can be applied when treatments are assigned based on a variable crossing a threshold \citep{thistlethwaite1960regression, imbens2008regression}. In medicine, thresholds defined by clinical parameters frequently serve as decision rules for treatment assignment. Despite their clinical utility, such thresholds create an artificial boundary: patients just above typically receive treatment, while those just below do not. Since individuals near the threshold are highly similar in observed and unobserved characteristics but differ minimally in the threshold-defining variable (and thus treatment exposure), RDD can estimate the causal effect of the intervention on outcomes of interest.

\subsection{Threshold-based decision-making in medicine}
\begin{figure}[ht]
    \centering
    \includegraphics[width=0.65\textwidth]{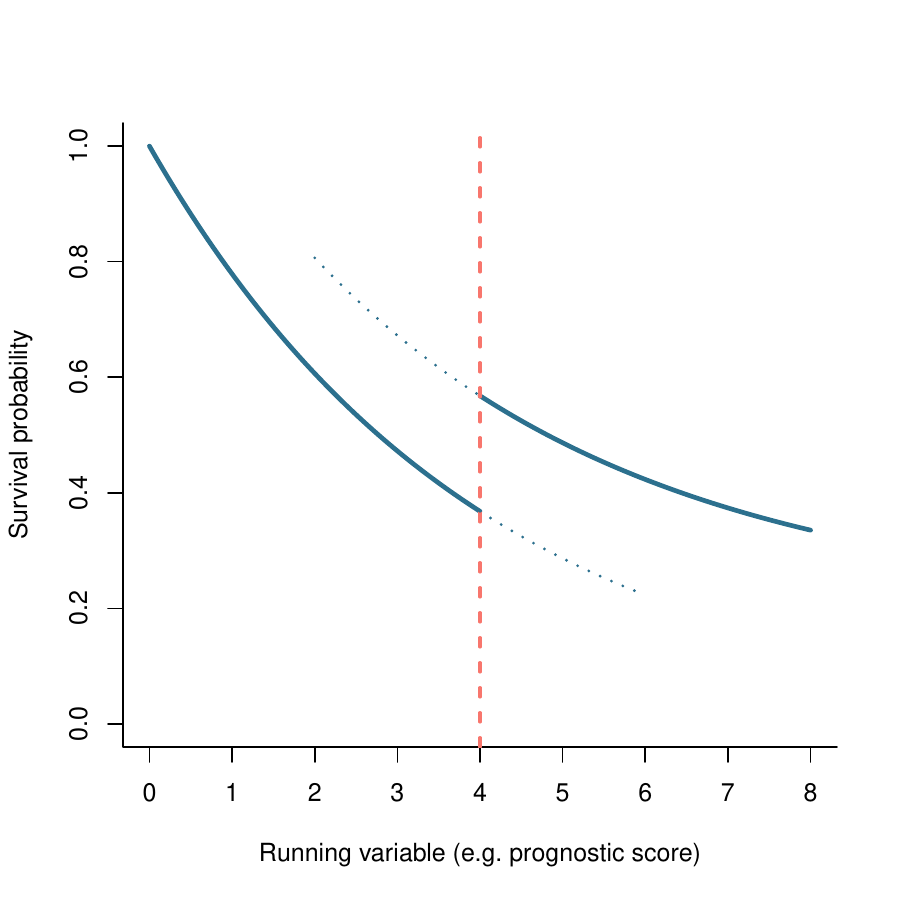}
    \caption{Illustration of the canonical RD design. Survival is declining as the prognostic running variable increases. At the threshold (dashed line) an intervention is triggered, resulting in a substantial improvement in the prognosis.}
    \label{fig:rd_canonical}
\end{figure}

The ubiquity of thresholds in medicine makes RDD a highly applicable study design. Figure \ref{fig:rd_canonical} gives a stylized example of settings in which some prognostic variable serves as a decision rule for treatment. For example, age thresholds often determine who is eligible for a screening or prevention policy \citep{eyting2025natural}. In general medicine, diabetes and hypertension diagnoses and treatment are based on HbA1c \citep{elsayed20242, lemp2023quasi} and systolic blood pressure \citep{AHA_guideline}, respectively. In oncology, tumor-specific biomarkers often determine who should receive an additional therapy \citep{herbst2020atezolizumab}. The widespread implementation of thresholds in clinical guidelines makes RDD a valuable quasi-experimental design to complement causal evidence from randomized controlled trials. This is particularly useful when trials do not reflect real-world patient populations or the exact clinical decisions that clinicians face in routine care \citep{degtiar2023review}. Additionally, RDDs can help evaluate thresholds that were established based on weak evidence or exploratory analyses. When clinical trials are practically or ethically infeasible, or when their results are inconclusive, quasi-experimental designs like RDD may then provide the best available evidence.

The prostate component of the Prostate, Lung, Colorectal, and Ovarian (PLCO) Cancer Screening Trial (hereafter referred to as the PLCO trial) is a prime example of how an RDD can provide additional clinical evidence when results from a randomized trial are inconclusive. Prostate cancer is among the most common types of cancer, with approximately 1.5 million new cases and 400,000 deaths per year globally \citep{bray2024global}. Following a series of randomized controlled trials, the U.S. screening guidelines recommend that prostate cancer screening for men be determined on a case-by-case basis through shared decision-making between patient and clinician \citep{grossman2018screening}. The PLCO trial investigated whether prostate cancer screening, consisting of a blood-based prostate specific antigen (PSA) test and a physical exam, offered a survival benefit compared to no screening \citep{andriole2009mortality}. The results from the trial remained inconclusive and controversial due to high rates of contamination in the control arm: while 50\% of participants were randomized to treatment, the majority of patients in the control group reported having undergone at least one PSA test. To overcome this limitation, \citet{shoag2015efficacy} propose a causal design that focuses on the receipt of prostate biopsy as part of the screening cascade in the treatment arm of the PLCO trial. Patients were eligible for this diagnostic surgical procedure if their PSA measurement exceeded 4.0 ng/mL. Patients at and below this threshold were not, resulting in a feasible regression discontinuity design. 

\subsection{Time-to-event outcomes}
A particular statistical challenge in the PLCO trial and RDD in healthcare more broadly is the presence of right-censoring in settings with time-to-event outcomes. To assess the benefit of an intervention, practitioners are frequently interested in (long-term) survival endpoints such as overall survival, disease-specific survival, or time until a disease onset/progression. The disease context of this study, prostate cancer, is often indolent, with a prolonged latency before events such as death or metastatic spread occur. As a result, right-censoring due to drop-out or loss to follow-up presents an important statistical challenge and its adjustment is highly relevant for the unbiasedness and precision of causal estimates.

Traditionally, the most common approach to account for right-censoring is to use inverse probability of censoring weighting (IPCW, \citet{laan2003unified}). This approach weighs the observed events by their inverse probability of remaining in the study, i.e., being censored beyond a given time point, and discards the censored observations. RDDs are sensitive to sample size, so discarding observations can cause a substantial loss in power. Moreover, the IPCW approach is also prone to bias if the estimated censoring probabilities are inaccurate.

In the context of RDDs specifically, \citet{adeleke2022regression} suggest using accelerated failure models to address right-censoring. This approach replaces the regression models typically employed for estimating the effect at the threshold with time-to-event models that account for censoring.  A drawback of this approach is that the regression model does not incorporate covariates. \citet{cho2021analysis} improve on this approach using doubly robust censoring corrections and pair these with the inferential approach of \citet{calonico2014robust}. This approach leverages both censoring and event models and thus yields unbiased results if either model is correctly specified \citep{robins1994estimation,tsiatis2006semiparametric, laan2003unified}. A limitation of this solution is that it still depends on parametric assumptions in the working model for the survival and censoring distributions. 

A more credible approach to survival analysis is to rely on flexible nonparametric machine learning methods that can incorporate information from patient-level covariates \citep{van2011super, westling2024inference}. This is relevant when a given study population is composed of several subgroups that exhibit covariate-specific survival and censoring patterns. Here, we extend this perspective to RDDs by estimating the nuisance components of the doubly robust censoring corrections using flexible machine learning models \citep{dml, van2011super}. This strategy leverages existing estimation and inference tools for RDDs (e.g., \citet{armstrong2018optimal, calonico2014robust, ghosh2025plrd}), but replaces the censored outcomes with pseudo-outcomes estimated from flexible nonparametric models. The use of pseudo-outcomes in place of unobserved variables has a rich history in both causal inference (e.g., \citet{kennedy2023towards}) and survival analysis \citep{andersen2017causal, cho2021analysis, su2022causal}. We demonstrate this approach in a detailed re-analysis of the prostate component of the Prostate, Lung, Colorectal, and Ovarian (PLCO) Cancer Screening Trial \citep{andriole2009mortality, prorok2000design}, suggesting that a nonparametric method may offer complementary insights and potentially more credible causal conclusions. 
Our analysis of the PLCO trial includes a large and heterogeneous cohort, where censoring patterns arecomplex and vary by the trial participants' demographics and many other descriptors. This offers a rich application as real-world data such as electronic health records are often subject to considerable fluctuations by irregular data entries and updates, drift in the distribution of covariates and/or outcomes, or external shocks. In the presence of such variation, parametric assumptions are frequently violated, and causal estimates are prone to bias. This is further complicated by the fact that in real-world observational settings, where RDDs are most appealing, censoring rates are often substantially higher than in randomized controlled trials. 

An open source software package, \texttt{rdsurvival} for \texttt{R} \citep{Rcore}, implements the proposed method.

\section{Regression discontinuity designs with survival outcomes} \label{sec:rd_survival}
For every unit $i$, we have access to a running variable $Z_i \in \mathbb{R}$ that determines treatment eligibility via a certain threshold $c \in \mathbb{R}$. In a ``sharp'' RD setting, the treatment $W_i \in \{0, 1\}$ is deterministically assigned for every unit for which the running variable exceeds this threshold, i.e., $W_i = 1(Z_i \geq c)$. We are interested in the effect of a given treatment on a time-to-event $T_i$, e.g., the time from treatment assignment until death.

By limiting our survival analysis up to a time horizon $h$, we are able to construct several effect measures. Two simple and interpretable estimands are the effect on i) $h$-month survival probability and ii) the impact on mean $h$-month survival, where $0 < h < \infty$ is some  chosen finite time point\footnote{These estimands are simple to interpret and estimate, as they are linear in the distribution of $T_i$. For more complex survival estimands in the RD setting, such as hazard ratios, see, e.g., \citet{stoltenberg2022regression}.}. For example, in a study where we have outcome data for up to 60 months following treatment assignment, we may determine an $h$ up to 60 months based on domain expertise. An alternative to pre-specified horizons is to look at a feasible time point in a data-driven way, for example, by setting $h$ to the 90-th percentile of observed follow-up times \citep{tian2020empirical}. 

Let $T_i(1)$ and $T_i(0)$ denote the potential outcomes \citep{imbens2015causal} in the two treatment states for the $i$-th unit in the presence of control and treatment, respectively. For survival probabilities, the RD setting allows us to identify the following counterfactual difference for the population with a running variable $Z_i$ equal to the threshold $c$.

\begin{defi}\label{def:tau}
    The RD treatment effect on time-$h$ survival probability is
    \begin{equation} \label{eq:deftau}
    \taurisk = \PP{T_i(1) > h \mid Z_i = c} - \PP{T_i(0) > h \mid Z_i = c}.
    \end{equation}
\end{defi}

\begin{figure}[t]
    \centering
    \includegraphics[width=0.65\textwidth]{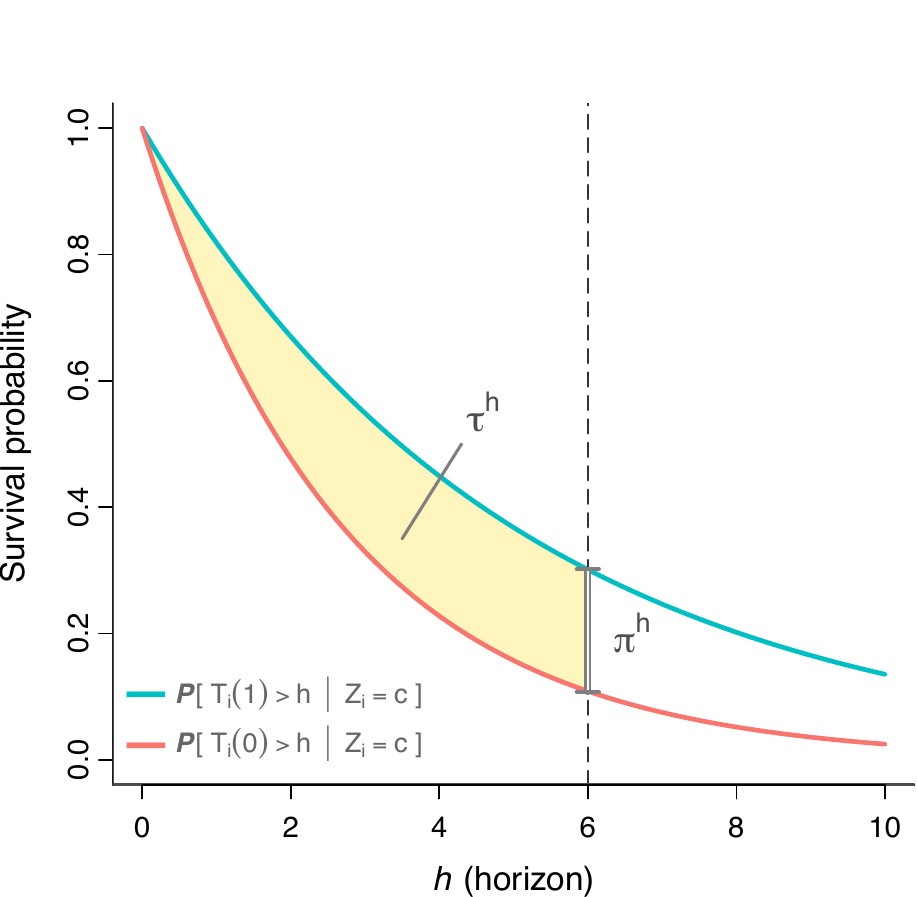}
    \caption{Illustration of the estimands $\taurisk$ and $\taurms$. For the population with running variable $Z_i$ equal to the threshold $c$, $\taurisk$ is the vertical distance between to counterfactual survival curves $\PP{T_i(w) > h \mid Z_i = c}$, $w \in \{0,1\},$ at time point $h$, while $\taurms$ is the area between the counterfactual survival curves up to time point $h$.}
    \label{fig:rd_estimand}
\end{figure}

The second estimand we consider is the effect on mean time-$h$ survival. A purely mechanical issue we have to deal with when estimating mean survival times is the issue of truncation. In a study with a follow-up time of 60 months, we barely observe any patients past 60 months, and so it is reasonable to ask for the effect on at most mean 60-month survival. Truncated means formed this way are often referred to as restricted mean survival times (``RMST''; see e.g., \citet{royston2013restricted}). 
\begin{defi}\label{def:tau_rmst}
    The RD treatment effect on mean time-$h$ survival (``RMST'') is
    \begin{equation} \label{eq:deftau_rmst}
    \taurms = \EE{\min(T_i(1), h) \mid Z_i = c} - \EE{\min(T_i(0), h) \mid Z_i = c}.
    \end{equation}
\end{defi}
A convenient property of survival functions is that the mean time-$h$ survival is equal to the area under the survival curve up to time $h$, and so \eqref{eq:deftau_rmst} can be seen as an effect measure that aggregates \eqref{eq:deftau}.
The practical interpretation of these quantities is straight-forward, for example, in a medical setting where the survival outcome is cancer-free
 survival, and the running variable determines a treatment, then for $h=24$, \eqref{eq:deftau} quantifies the difference in survival probability past 2 years when receiving biopsy, and \eqref{eq:deftau_rmst} quantifies the differential effect on mean 2-year survival when receiving a biopsy. Figure \ref{fig:rd_estimand} gives a graphical representation of these quantities.

\subsection{Estimation with observed outcomes}\label{sec:complete_estimation}
At first glance, it might be tempting to view the RD design as a special version of a localized randomized controlled trial where it suffices to compare mean outcomes on both sides near the threshold. From an estimation perspective, this overlooks a core insight in many RDDs: the running variable (e.g., disease severity) is highly prognostic of the outcome (e.g., survival). In order to disentangle the gradual effect of the running variable on the outcome from the treatment effect, it is common to assume the conditional mean functions in \eqref{eq:deftau} and \eqref{eq:deftau_rmst} are continuous and that the running variable has continuous support around the threshold \citep{hahn2001identification, imbens2008regression}. $\taurisk$ is then identified by the following limit in terms of the observed time-to-event $T_i$
$$
\taurisk = \lim_{z \downarrow c} \EE{1(T_i > h) \mid Z_i = z} - \lim_{z \uparrow c} \EE{1(T_i > h) \mid Z_i = z},
$$
where we've used the simple fact that $\PP{T_i > h} = \EE{1(T_i > h)}$. The second estimand $\taurms$ is identified by
$$
\taurms = \lim_{z \downarrow c} \EE{\min(T_i, h) \mid Z_i = z} - \lim_{z \uparrow c} \EE{\min(T_i, h) \mid Z_i = z}.
$$
These limits are commonly estimated using local linear regressions on each side of the discontinuity threshold $c$ \citep{hahn2001identification}, or, as advocated by \citet{imbens2019optimized}, through minimax optimal convex optimization. Given access to survival data with observed outcomes, it is straightforward to obtain estimates and confidence intervals of the parameters $\taurisk$ or $\taurms$: simply call into a continuity-based RD estimator of our choice (e.g., \texttt{rdrobust} \citep{rdrobust}, \texttt{plrd} \citep{ghosh2025plrd}, \texttt{RDHonest} \citep{rdhonest}) with outcome equal to the argument in the expectation operator for the respective estimand: $1(T_i > h)$ for \eqref{eq:deftau} and $\min(T_i, h)$ for \eqref{eq:deftau_rmst}.

\subsubsection{Fuzzy RDs}\label{sec:fuzzy_rd}
These previous estimands and their interpretation naturally carry over to the ``fuzzy RD'' setting, where units above the treatment threshold $c$ are more likely to receive treatment than those units below the threshold. Here, the treatment assignment $W_i$ is not equal to $1(Z_i \geq c)$, but rather the treatment probability $\PP{W_i = 1 \mid Z_i}$ jumps at the value $Z_i = c$.  Depending on the application, the estimands $\taurisk$ and $\taurms$ can be interpreted as simple intention-to-treat (ITT) parameters, measuring the effect of being \emph{assigned} treatment, but not necessarily \emph{receiving} it. Estimating these ITT parameters simply amounts to proceeding as in the previous paragraph, treating the estimation problem as a sharp RD, but modifying the interpretation of the result. 

We can also estimate local RD parameters that measure the effect of receiving treatment by invoking assumptions similar to those used in instrumental variables settings with imperfect compliance \citep{imbens2008regression}. The interpretation of $\taurisk$ is then the same, but restricted to the compliers only (i.e., those units who receive treatment exactly when assigned treatment). We denote the jump in treatment probability by
$p = \lim_{z \downarrow c} \EE{W_i \mid Z_i = z} - \lim_{z \uparrow c} \EE{W_i \mid Z_i = z}$. This local fuzzy counterpart to the estimand \eqref{eq:deftau} is then identified by the following simple rescaling
$$
\widetilde \pi^{h} = \frac{\taurisk}{p}.
$$
The fuzzy RD estimand is not identified when the jump in treatment probability $p$ is 0. It is also unstable when this probability difference is close to 0, highlighting an important role of the design stage when constructing fuzzy RD research designs. Section \ref{sec:app_prostate} gives an example of this design step by carefully defining the target population of eligible patients. \citet{feir2016weak} and \citet{noack2024bias} suggest Anderson-Rubin-type tests for conducting inference on local RD parameters that can be used in settings where $p$ is close to zero.

For further discussion of RD estimation and inference see \citet{cattaneo2024practical}, \citet{kolesar2024eco539b}, and \citet[Ch. 8]{wagerbook}.

\subsection{Estimation with right-censored outcomes}
A challenge with many studies on survival outcomes is that we do not observe $T_i$ for every unit $i$. Due to drop-out or loss to follow-up, the event time $T_i$ is not observed for all units; instead we observe the time at which a unit is censored $C_i$.
\begin{defi} \label{def:censoring}
Under right-censoring, we observe $Y_i = \min(T_i, C_i)$ where $T_i$ is the survival time and $C_i$ the censoring time, along with a non-censoring indicator $\Delta_i = 1(T_i < C_i)$.
\end{defi}

\begin{figure}[t]
    \centering
    \includegraphics[width=1 \textwidth]{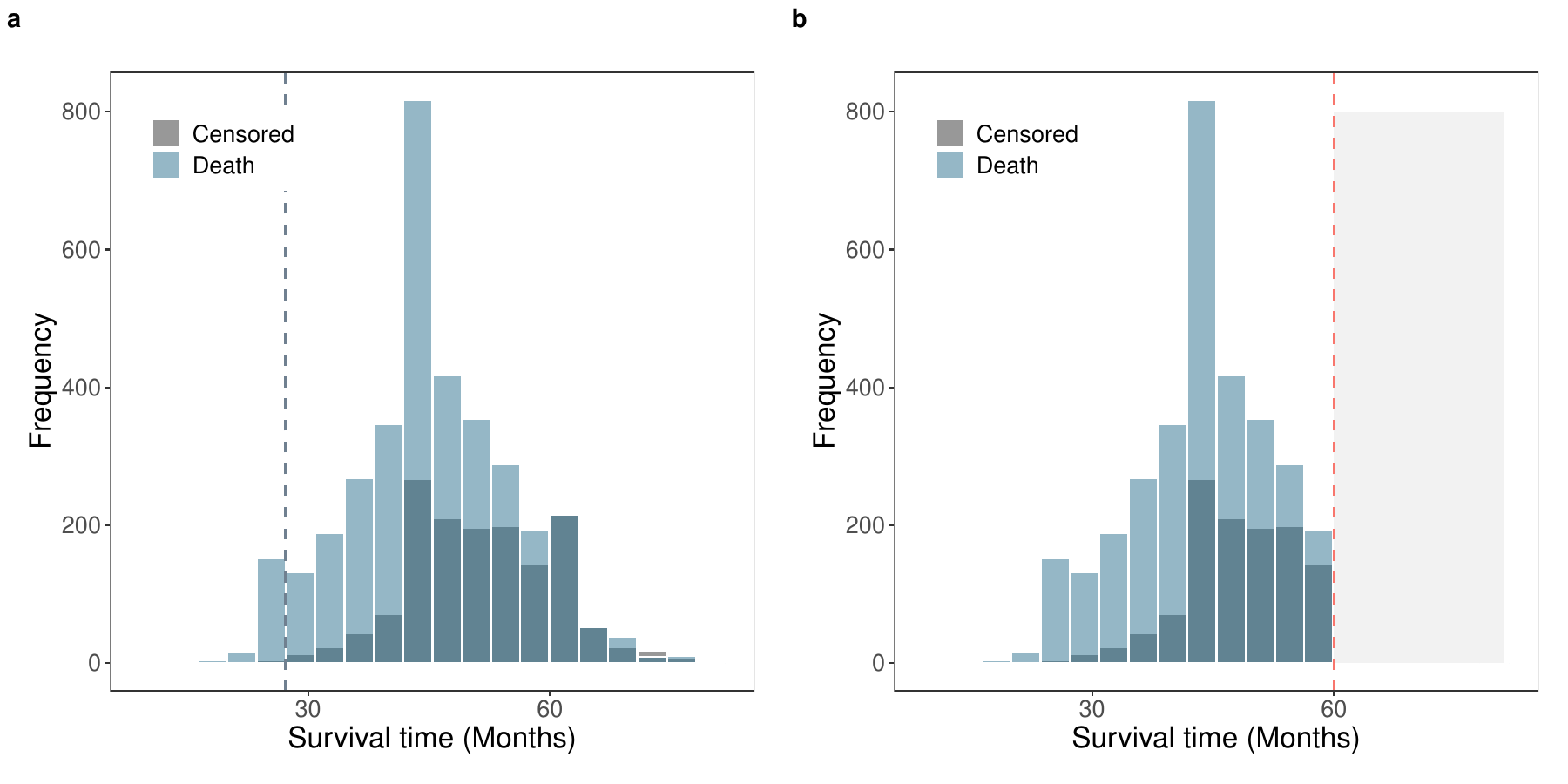}
    \caption[]
    {The figure shows two scenarios with a total of around 70 months of follow-up. In the first scenario (panel a), censoring poses no challenge up to 28 months (dashed gray line). In the second scenario (panel b), estimating the effect on $h = 60$ months (dashed red line) involves remapping censoring and event times observed past $h$.}
    \label{fig:histogram1}
\end{figure}

This right-censoring pattern is ubiquitous in applications that use electronic medical records, registry data, or studies where patients are enrolled at different time points. Figure \ref{fig:histogram1} illustrates what the data may look like in a hypothetical study with 60 months of follow-up time. Up until $h=28$ we observe only events (e.g. deaths) for subjects in the study. Had we only been interested in the effect on survival up to 28 months, then this situation would pose no additional complications for us, since it is only \emph{after} this time point that units are censored. Estimating an $h = 28$-month RD effect here only involves remapping every unit we tracked past $h$ to $h$ and treating them as observed: if we know they were censored in the future, we know they must have been alive at $h$.

This situation becomes more challenging when asked about the effect of survival past $h = 30$ months. If we are interested in the effect on $h=60$ month (5-year) survival, for example, then there is a considerable number of patients that are censored. In Figure \ref{fig:histogram1}b, we have drawn a vertical line at which we remap everyone tracked past this point. Now, how do we address the estimation of \eqref{eq:deftau} and \eqref{eq:deftau_rmst} in this scenario? Simply ignoring these samples would in general lead to biased estimates. In order to obtain unbiased estimates, we need to account for the missingness pattern imposed by censoring. In order to do so, we first need to make some assumptions about the censoring mechanisms. To accommodate realistic clinical settings, we assume access to a set of $d$ time-invariant patient characteristics $X_i \in \mathbb{R}^d$ observed at baseline. These could be clinical variables, such as disease severity (e.g. mitotic rate of a tumor), or demographics such as age, that we believe can be predictive of both survival and censoring.

\begin{assu} \label{asu:ignorable}
    (Conditionally ignorable censoring). Conditional on covariates and the running variable, censoring is ignorable,
    $$
    T_i \indep C_i \cond X_i, Z_i.
    $$
\end{assu}
This ignorable censoring assumption is strong and non-testable, but standard in the literature \citep{fleming2013counting}. Note that, by conditioning on $X_i$, we allow for the censoring structure to vary according to some observable predictor variables, which alleviates a concern that not every patient is equally likely to be censored. Next, we need a more mechanical assumption that relates our chosen time point of interest $h$ to the censoring process. The following assumption ensures that at the given time point $h$, censoring has not completely masked all the clinically relevant information we care about.
\begin{assu} \label{asu:positivity}
    (Positivity). Some patients are never censored, there is an $\eta > 0$ such that for our horizon $h$ of interest,
    $$
    \PP{C_i > h \mid X_i = x, Z_i = z} \geq \eta ~\text{for all}~ x ~\text{and}~ z.
    $$
\end{assu}
Note that these assumptions are essentially counterparts to the \emph{unconfoundedness} and \emph{overlap} assumptions in an observational study \citep{rosenbaum1983central}. The positivity assumption is typically easier to accommodate in survival settings since we can  often pick a time horizon where positivity is not an issue and still obtain an interpretable estimand. In Figure \ref{fig:histogram1}, positivity is clearly an issue beyond $h=65$ months due to considerable censoring and few observed events. Lowering the horizon to $h=60$ yields an interpretable 60-month survival estimate that alleviates this concern.

\begin{rema}
    In settings where $T_i$ measures, for example, time until cancer progression, a competing event is something that precludes this event from happening, e.g., death from a heart attack. In these settings, the researcher can choose to interpret $\taurisk$ and $\taurms$ as either direct or total effects \citep{young2020causal}, depending on whether competing events are defined as censoring events. See \citet{rubin2006causal} for further conceptualizations of causal effects in settings with competing events.
\end{rema}

\subsection{Accounting for censoring via weighting}
Given these assumptions, it is conceptually straightforward to account for censoring with weighting. Given Definition \ref{def:censoring}, the probability we observe a unit (i.e., the unit is not censored) is
$$
\PP{\Delta_i = 1 \mid X_i, Z_i} = \PP{C_i > T_i \mid X_i, Z_i}.
$$
The idea of inverse probability of censoring weighting (IPCW) is to estimate \eqref{eq:deftau} using the samples for which we observe the event time $T_i$, but up-weight them with weights equal to the inverse of the observation probabilities \citep{laan2003unified}. Assumption \ref{asu:positivity} on positivity ensures that these inverse weights do not explode. The following procedure outlines the steps (for ease of exposition, our focus will be on discussing the underpinnings of estimating $\taurisk$ , but note that the approach readily extends to {$\taurms~$}\footnote{Or any deterministic linear transformation of $T_i$ bounded by a finite horizon $h$. The \texttt{R} package \rdsurvival supports both $\taurisk$ and $\taurms$.}).

\begin{alg} \label{alg:IPC}
    IPC-adjusted RD estimate for $\taurisk$.
\end{alg}
\begin{itemize}
    \item[1.] Form estimates $\widehat S_C(t)$ of the censoring process. This can be done by fitting a Kaplan-Meier curve using $Y_i$ as outcomes and $1 - \Delta_i$ as the censoring indicator.
    \item[2.] Construct sample weights equal to $w_i = 1 / \widehat S_C(\min(h, Y_i))$.
    \item[3.] For the subset of samples with $\Delta_i=1$ or that are censored after $h$, call into an RD estimator with outcomes equal to $1(Y_i > h)$ and sample weights equal to $w_i$. Return the estimated RD parameter and confidence intervals.
\end{itemize}

\begin{figure}[t]
        \centering
        \includegraphics[width=0.9\textwidth]{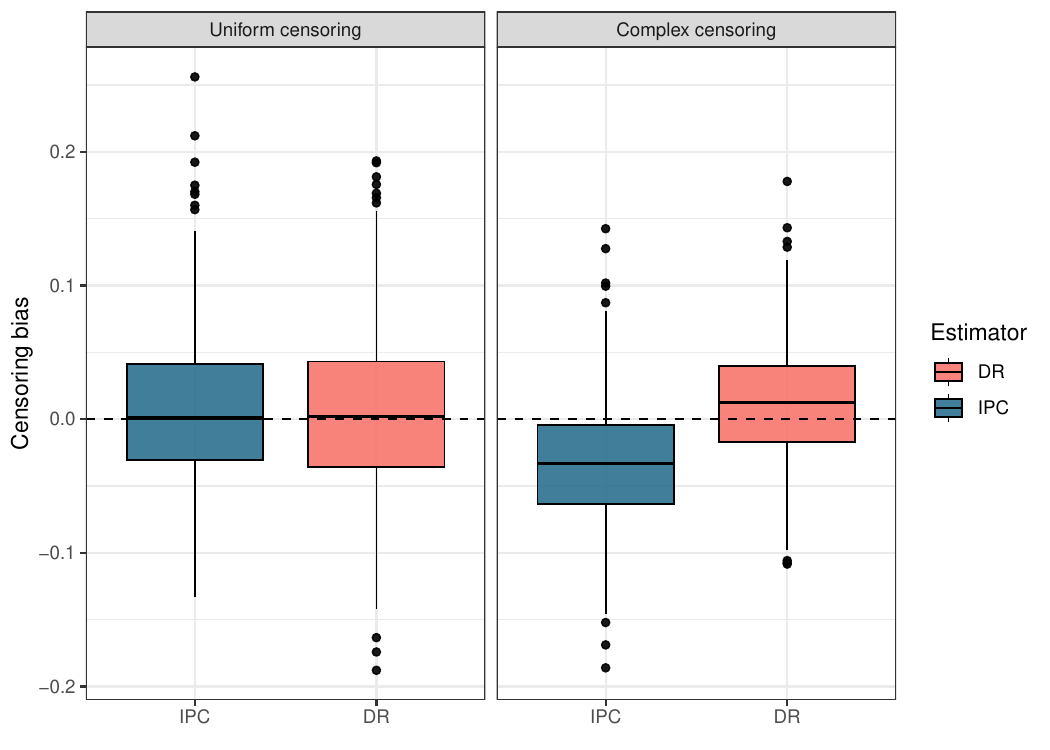}
        \caption[]
        {Example of bias using an IPCW approach compared with a more flexible doubly robust (DR) approach that uses nonparametric survival estimators, using simulation settings described in Section \ref{sec:simulation} (uniform censoring is ``setting 1'' and complicated censoring is ``setting 2'') and a sample size of $n=5,000$. The boxplots show censoring bias defined as the difference between an estimate $\hat \pi^{h}$ using censor-corrections, and an estimate $\hat \pi_*^{h}$ using the uncensored complete-data $T_i$, from the same realization of $n=5,000$ observations.}
        \label{fig:bias_bosplot}
\end{figure}

The drawback of this approach is that we have to estimate the censoring probabilities, and the quality of these estimates completely determines our ability to successfully account for censoring. A workhorse approach for this task is survival curves with Kaplan-Meier estimators. These, however, cannot easily account for possibly complicated censoring mechanisms that vary by patient characteristics, which can be highly relevant, as our conditioning on $X_i$ and $Z_i$ implies. Figure \ref{fig:bias_bosplot} illustrates these complications by showing estimated bias arising from censoring for IPCW-adjusted RD estimates and compares them with a more flexible approach (DR) described in the next section.

On the left-hand side of Figure \ref{fig:bias_bosplot}, the IPCW approach performs well because the simulated censoring pattern is simple and follows a uniform distribution. This is something an unconditional Kaplan-Meier curve has little problem with estimating correctly. In the other simulation setting, the censoring pattern is more complicated, and the IPCW approach fails to successfully eliminate censoring effects. A second drawback is that in Step 3 of Algorithm \ref{alg:IPC} we are discarding all samples that are censored before $h$. Since RD estimation is data-intensive, reducing the number of observations can result in a significant loss of power. The doubly robust approach (DR), performs better by relying on flexible machine learning approaches to estimate survival curves conditional on $(X_i, Z_i)$, which are used to construct ``pseudo-outcomes'' \citep{andersen2017causal, steingrimsson2019censoring} that take the place of the unknown complete-data outcomes $T_i$ in the RD estimation. The next section elaborates on this strategy.

\begin{rema}
In Step 3 of Algorithm \ref{alg:IPC}, we are effectively treating units that are tracked past $h$ but eventually censored to be observed for the purpose of estimating $\taurisk$. This is simply a consequence of Definition \ref{def:tau} and \ref{def:tau_rmst} in terms of a finite time point $h$: anything that happens after this time point is not relevant to us. A good practice in analyzing censored survival data is to plot a histogram of events by censoring and event status to assess how much censoring is affecting our estimand. If no censoring occurs before the time point of interest $h$, then there is no need to employ censoring corrections. 
\end{rema}

\subsection{Doubly robust censoring corrections}
The general problem in the preceding section is that the goal of consistent RD estimation hinges entirely on a single model to correctly specify and account for censoring. In real-world scenarios, it is often very hard or impossible to know what the underlying censoring process looks like. A key point to recognize in the previous section is its conceptual relation to estimating an average treatment effect under unconfoundedness, where we can perform the confounding adjustment via either weighting (using a propensity score), a regression adjustment (using a conditional mean function), or a doubly robust combination of both \citep{robins1994estimation}. This doubly robust combination typically enjoys superior properties over using either approach alone \citep{bang2005doubly, ding2023first, wagerbook}.   
An exact analog for this doubly robust estimator to survival settings with censoring is via censoring robust transformations \citep{rubin2007doubly, tsiatis2006semiparametric}, where, in addition to estimates of the censoring process used for weighting, we utilize a regression counterpart in the form of estimates of the survival curve $\PP{T_i > t \mid X_i, Z_i}$. 

Given a continuity assumption on the censoring distribution, it is possible to construct generic censoring robust transformations for a large class of estimation problems (e.g., \citet{tsiatis2006semiparametric}) that augment complete-data formulations with doubly robust censoring corrections\footnote{Given the complete-data estimating equation $\psi_\theta^c(T_i)=T_i - \theta$ for a target parameter $\theta$ depending on the survival time $T_i$, a censoring-augmented estimating equation is given by the well-known transformation $\Delta_i \psi_\theta^c(T_i)/S_C(Y_i) + (1-\Delta_i)\EE{\psi_\theta^c(T_i) \mid T_i > Y_i}/S_C(Y_i) - \int_{}^{Y_i} dS_C(t)/S^2_C(t) \EE{\psi_\theta^c(T_i) \mid T_i > t} dt$.}. This is the approach \citet{cui2023estimating} takes to estimate conditional average treatment effects under right-censoring, and \citet{cho2021analysis} in estimating censoring doubly robust RDs (using parametric models for the censoring and survival process). As is often the case, working with continuous-time censoring is largely a mathematical convenience, but does not reflect the medical or administrative reality of the data. In our practical applications, censoring arises due to various data collection protocols (e.g., scheduled follow-up for registry data), electronic health record administration (e.g., irregular data pulls), or unexpected events (e.g., patients moving to another state). Rather than happening at a continuous rate, censoring processes take place at varying discrete steps during a fixed study period. Approaching censoring processes via covariate-conditional Kaplan-Meier curves on a continuous time grid is, therefore, problematic.

Here, instead of relying on continuous-time censoring unbiased transformations, we use a version that accounts for discrete-time censoring. We assume that follow-up with individuals happens at pre-determined times $t=\{1, \ldots, t_{max}\}$ and that we know the status of each patient. That is, we are able to ascertain if a patient is either alive or censored--and we are thus unable to observe the complete time-to-event due to loss to follow-up. We assume each unit $i$ has a censoring time $C_i = \{1, \ldots, +\infty\}$, where $+\infty$ denotes that the patient is never censored. Under this setup, then for the RD estimand \eqref{eq:deftau}, the goal is to derive a censoring doubly robust estimator for $\PP{T_i > h} = \EE{1(T_i > h)}$. To do so, we can leverage a doubly robust estimator for a discrete-time dynamic policy \citep[Ch. 14]{wagerbook}. The idea involves recognizing that estimating $\PP{T_i > h}$ is the same as estimating the value of a certain dynamic policy, where we appropriately interpret policy values as survival probabilities and propensity scores as censoring probabilities. For details, we refer to \citep[Exercise 14]{wagerbook}, and here state the resulting doubly robust construction\footnote{Taking the limit as $t \rightarrow t+1$ in \eqref{eq:drscores} yields the familiar continuous-time expressions in \citet{cui2023estimating, rubin2007doubly, xu2023treatment}, which require estimating an integral involving the censoring hazard and censoring process. In contrast, \eqref{eq:drscores} avoids this and remains consistent even with coarse time grids, where continuous-time estimators can suffer from discretization error (e.g., monthly time steps over a short follow-up).}.

Let $S_T(t; x, z) = \PP{T_i > t \mid X_i = x, Z_i = z}$ and $S_C(t; x, z) = \PP{C_i > t \mid X_i = x, Z_i = z}$ denote the survival curves for the survival and censoring process, respectively. Define $H_i = \min(Y_i, h)$, and the following score
\begin{align} \label{eq:drscores}
 \widehat \Gamma_i &= \widehat S_T(h; X_i, Z_i) + \\
 &\sum_{t=1}^{H_i - 1} \frac{1}{\widehat S_C(t; X_i, Z_i)}
  \left(\frac{\widehat S_T(h; X_i, Z_i)}{\widehat S_T(t; X_i, Z_i)} - \frac{\widehat S_T(h; X_i, Z_i)}{\widehat S_T(t - 1; X_i, Z_i)}\right) +\nonumber \\
&\frac{\max(\Delta_i, 1(Y_i > h))}{\widehat S_C(H_i; X_i, Z_i)}
  \left(1(Y_i > h) -  \frac{\widehat S_T(h; X_i, Z_i)}{\widehat S_T(H_i - 1; X_i, Z_i)}\right). \nonumber
\end{align}
If the curves $\widehat S_T(t; X_i, Z_i)$ and $\widehat S_C(t; X_i, Z_i)$ are estimated sufficiently well, then  $\sum_{i=1}^{n} \widehat \Gamma_i$ is an efficient doubly robust estimate of $\PP{T_i > h}$\footnote{A similar doubly robust construction readily follows for $\EE{\min(T_i, h)}$ required for the RMST estimand \eqref{eq:deftau_rmst}; here, estimates of $\EE{\min(T_i, h) \mid T_i > t, X_i, Z_i}$ replaces terms of the form $\widehat S_T(h; X_i, Z_i) / \widehat S_T(t; X_i, Z_i) := \PP{T_i > h \mid T_i >t, X_i, Z_i}$. These can be estimated via the area under the estimated survival function $\widehat S_T(t; x, z)$.}.
We can motivate debiased machine learning arguments for estimating $\widehat S_T(t; X_i, Z_i)$ and $\widehat S_C(t; X_i, Z_i)$ by forming the score $\widehat \Gamma_i$ without using the $i$-th unit for estimating the survival or censoring process. This can be done using suitable cross-fold or sample-splitting constructions, or in the case of forests, out-of-bag predictions. This is the approach \rdsurvival takes, by forming estimates using random survival forests \citep{ishwaranRSF} as implemented in \texttt{grf} \citep{GRF}. This effectively leverages random forests to deliver flexible and data-driven Kaplan-Meier (or Nelson-Aalen) curves that are conditional on ($X_i, Z_i$) and gives us the following approach to flexibly account for censoring.

\begin{alg} \label{alg:DRrdd}
    Doubly robust censoring adjusted RD estimate for $\taurisk$.
\end{alg}
\begin{itemize}
    \item[1.] Estimate the survival and censoring processes $\widehat S_T(t; X_i, Z_i)$ and $\widehat S_C(t; X_i, Z_i)$ using flexible survival estimators, such as random survival forests.
    \item[2.] Construct doubly robust scores $\widehat \Gamma_i$ given by \eqref{eq:drscores} using out-of-bag estimates.
    \item[3.] Call into an RD estimator using $\widehat \Gamma_i$ as outcomes and return estimate and confidence interval.
\end{itemize}

The type of guarantees we can obtain with this doubly robust approach is that in Step 3 of Algorithm \ref{alg:DRrdd}, by using outcomes $\widehat \Gamma_i$ we are doing asymptotically as well as if we had run Step 3 using the (infeasible) complete data outcomes $1(T_i > h)$, i.e., we obtain confidence intervals that are valid for the given RD estimator. The formalization of this result follows from the approach in Armstrong (2020).

\section{Simulation study} \label{sec:simulation}
As indicated by Figure \ref{fig:bias_bosplot}, the doubly robust approach is more effective at eliminating censoring effects than IPCW. To assess the practical performance of our approach more thoroughly, we undertake a simulation study to evaluate the performance of the estimator implemented in \rdsurvival and verify that the doubly robust approach can reliably account for censoring across a range of scenarios. We consider four simulation settings where we generate $n$ independent pre-treatment covariates $X_i$ uniformly distributed on $[0, 1]^d$ with $d=10$, and the running variable $Z_i$ uniformly distributed on $[0, 1]$ with the treatment set to $W_i = 1(Z_i \geq 0.5)$. The survival time $T_i$ has a simple quadratic curvature in the running variable in all settings, but with varying degrees of censoring complications considered in \citet{cui2023estimating}.

In all settings, we consider estimating the effect of the treatment on both the survival probability and RMST at a given $h$. In \emph{Setting 1}, the survival time $T_i$ follows a Poisson distribution with mean
$X_1^2 + X_3 + 6 + 2(\sqrt{X_1} - 0.3) + Z_i^2 + 0.9W_i$ and the censoring time is uniformly distributed on $[1, 15]$. We consider a horizon parameter of $h=7$. In \emph{Setting 2}, the survival time $T_i$ follows the same Poisson distribution as in \emph{Setting 1}  with mean
$X_1^2 + X_3 + 6 + 2(\sqrt{X_1} - 0.3) + Z_i^2 + 0.9W_i$. The censoring process follows a Poisson distribution with mean
$10 + \log(1 + \exp({X_3})) - 1.5\cdot 1(Z_i \geq 0.5)$. We consider a horizon parameter of $h=9$. In \emph{Setting 3}, $T_i$ follows an accelerated failure time model with
$\log(T_i) = 1.8 + 0.7\sqrt{X_2} + 0.2X_3 - 0.4\sqrt{X_4} - 0.5 Z_i^2 + 0.75W_i + \epsilon_i
$, where $\epsilon_i \sim N(0, 1)$. The censoring process is uniform on $[0, 50]$, and we consider a horizon of $h=20$. Finally, in \emph{Setting 4} $T_i$ follows the same process as in \emph{Setting 3}, i.e.
$\log(T_i) = 1.8 + 0.7\sqrt{X_2} + 0.2X_3 - 0.4\sqrt{X_4} - 0.5 Z_i^2 + 0.75W_i + \epsilon_i
$. The censoring process follows a Cox model with parameters $\exp(-5.75 - 0.5\sqrt{X_2} + 0.2X_3 + 0.3\sqrt{X_4} Z_i)$ and baseline hazard $2t$. We consider a horizon of $h=15$. For the settings with continuous distributions, we record simulated event times with 1 digit and fit survival curves on the grid of all recorded times.

For each simulation setting, we use $n=5,000$ and $B=1,000$ Monte Carlo repetitions to estimate the survival probability and RMST at $h$, and calculate mean empirical coverage for $\taurisk$ and $\taurms$ using 95\% confidence intervals, average interval length, and root mean squared error (RMSE) $\sqrt{\frac{1}{B}\sum_{b=1}^{B}(\tau - \hat \tau_b)^2}$. The doubly robust censoring corrections are estimated using random survival forests, which is the default option in \texttt{rdsurvival}. For the RD estimator, we consider two approaches, \texttt{plrd} \citep{ghosh2025plrd} and \texttt{rdrobust} \citep{rdrobust}, using default options for both. These packages differ in their approach to constructing confidence intervals. \texttt{plrd} relies on ``bias-aware'' inference \citep{armstrong2018optimal}, while \texttt{rdrobust} relies on ``bias-adjusted'' inference \citep{calonico2014robust}. The results in Table \ref{tab:CI} show the coverage for both RD methods and survival estimands. In all settings, both RD methods achieve nominal coverage, with wider confidence intervals for \texttt{rdrobust} compared to  \texttt{plrd}. In RDDs, it is natural to believe that the running variable has a gradual effect on the outcomes, as Figure \ref{fig:rd_canonical} suggests. Estimators such as \texttt{plrd} utilize this structure to produce more efficient estimates (see \citet{armstrong2018optimal}, \citet[Ch. 8]{wagerbook} for an in-depth discussion). 

\begin{table}[t]
\thispagestyle{empty}
\centering
\begin{tabular}{lrrrrrr}
\multicolumn{7}{c}{\textbf{Operating Characteristics of DR Approach with Different RD Estimators \bigskip}}\\
\toprule
\multicolumn{1}{c}{Setting} & \multicolumn{3}{c}{Survival probability $\taurisk$} & \multicolumn{3}{c}{RMST $\taurms$} \\
\cmidrule(l{3pt}r{3pt}){2-4} \cmidrule(l{3pt}r{3pt}){5-7}
\multicolumn{1}{c}{} & \multicolumn{1}{c}{Coverage} & \multicolumn{1}{c}{RMSE} & \multicolumn{1}{c}{Length} & \multicolumn{1}{c}{Coverage} & \multicolumn{1}{c}{RMSE} & \multicolumn{1}{c}{Length} \\
\midrule
\multicolumn{7}{l}{\textit{\texttt{rdrobust}}}\\
\midrule
1 & 0.94 & 0.09 & 0.33 & 0.94 & 0.19 & 0.70 \\
2 & 0.95 & 0.08 & 0.29 & 0.95 & 0.26 & 1.01 \\
3 & 0.95 & 0.07 & 0.26 & 0.94 & 0.99 & 3.58 \\
4 & 0.95 & 0.08 & 0.28 & 0.95 & 0.69 & 2.61 \\
\midrule
\multicolumn{7}{l}{\textit{\texttt{PLRD}}}\\
\midrule
1 & 0.96 & 0.04 & 0.17 & 0.96 & 0.09 & 0.35 \\
2 & 0.94 & 0.04 & 0.15 & 0.96 & 0.14 & 0.51 \\
3 & 0.95 & 0.03 & 0.13 & 0.97 & 0.44 & 1.80 \\
4 & 0.95 & 0.04 & 0.14 & 0.95 & 0.35 & 1.31 \\
\bottomrule
\end{tabular}
\caption{Coverage of 95\% confidence intervals for $\taurisk$ and $\taurms$, RMSE, and average interval length for the four simulation settings described in Section \ref{sec:simulation}. The number of Monte Carlo repetitions is 1,000.}
\label{tab:CI}
\end{table}

\section{Survival Outcomes in the PLCO Cancer Screening Trial for Prostate Cancer} \label{sec:app_prostate}

To test our approach on a real data application, we consider the prostate cancer component of the U.S. Prostate, Lung, Colorectal, and Ovarian Cancer Screening Trial (PLCO) \citep{andriole2009mortality, prorok2000design}. This randomized controlled trial (RCT) was conducted from 1993 through 2001 and is among the most prominent studies aimed at assessing the benefits from prostate cancer screening \citep{andriole2009mortality, prorok2000design}. The trial included a total of 76,693 men aged 55 to 74 years at 10 U.S. centers who were randomized to either standard of care (control group) or annual screening (treatment group). The latter included annual prostate-specific antigen testing and digital rectal examination (DRE).

The PLCO trial provides an instructive example for how RDDs can complement the intention-to-treat (ITT) estimates from an RCT. Substantial contamination in the PLCO trial’s control arm diluted potential differences in outcomes and ultimately limited the interpretability of the trial. It turns out that RDD offers a practical workaround for inference: by restricting the analysis to the screening arm, one can exploit the threshold-based eligibility rules for downstream diagnostic procedures such as prostate biopsy. This is also the approach taken by \citet{cho2021analysis} and \citet{shoag2015efficacy}.

Specifically, we used the PSA-based eligibility threshold for prostate biopsy to isolate the effect of biopsy-based screening on survival outcomes in the screening arm ($n$=38,343). Prostate biopsy is a diagnostic surgical procedure to determine whether a patient has prostate cancer. In the PLCO trial, patients with PSA levels greater than 4 ng/mL were considered eligible for prostate biopsy. While patients above this threshold are, on average, at a higher risk for having prostate cancer \citep{american2000prostate}, the use of PSA remains controversial due to its limited specificity, moderate predictiveness of disease severity, and intra-individual variability. The potential for detecting and treating indolent prostate tumors that would never become clinically apparent during a patient's lifetime contributes to the screening controversy \citep{hamdy2023fifteen, wilt2012radical}. As a result, PSA-based thresholds for prostate biopsy are not strictly followed. The PLCO trial is no exception to that, resulting in a fuzzy RD design \citep{kasivisvanathan2018mri}. 
From a methodological perspective, the PLCO trial offers a compelling use case for survival analysis as it contains multiple time-to-event outcomes, including prostate cancer-free survival and overall survival, and follow-up times of 11.5 years \citep{andriole2009mortality} with post-trial follow-up. We define those outcomes as follows: prostate cancer-free survival is the number of days from randomization to the first occurrence of a cancer diagnosis, treating deaths and any loss to follow-up as censoring; overall survival (OS) is defined as the number of days from randomization to death from any cause. Since right-censoring can occur for both outcomes, the time-to-event is defined as the number of days from randomization to the event occurrence ($\Delta_i=1$), or to the last time the patient was confirmed to be event-free and alive ($\Delta_i=0$). Finally, the long-term follow-up motivates the use of pretreatment covariates in our statistical analysis as covariate-dependent survival differences are likely to emerge. To account for distributional differences, we included several pretreatment covariates, including age, marital status, occupation, smoking status, positive family history for prostate cancer, body mass index (BMI), and various comorbidities (e.g., diabetes) and medications (e.g,. aspirin). These are reasonable prognostic covariates to incorporate in our censoring and outcome models. For exampnlle, age and a positive family history for prostate cancer are known risk factors and might explain subgroup-specific survival differences \citep{garraway2024prostate, raychaudhuri2025prostate}. 

To prepare the data, we removed all individuals in the control group. We further excluded all individuals with a suspicious result in the DRE since those patients were eligible for a prostate biopsy irrespective of their PSA result. We further limited our analysis to the first measurement of PSA and prostate biopsy, which turned our analysis into a static RDD. In a second step, we excluded all individuals who had a missing value for the first PSA measurement, a personal history of prostate cancer, or had undergone at least one PSA test before enrollment. From this cohort, we selected all individuals with a PSA below 10 ng/mL, resulting in a final analytic cohort of $n = 17,351$ for our RDD.

\subsection{Data analysis}
\begin{figure}[t]
        \centering
        \includegraphics[width=0.7\textwidth]{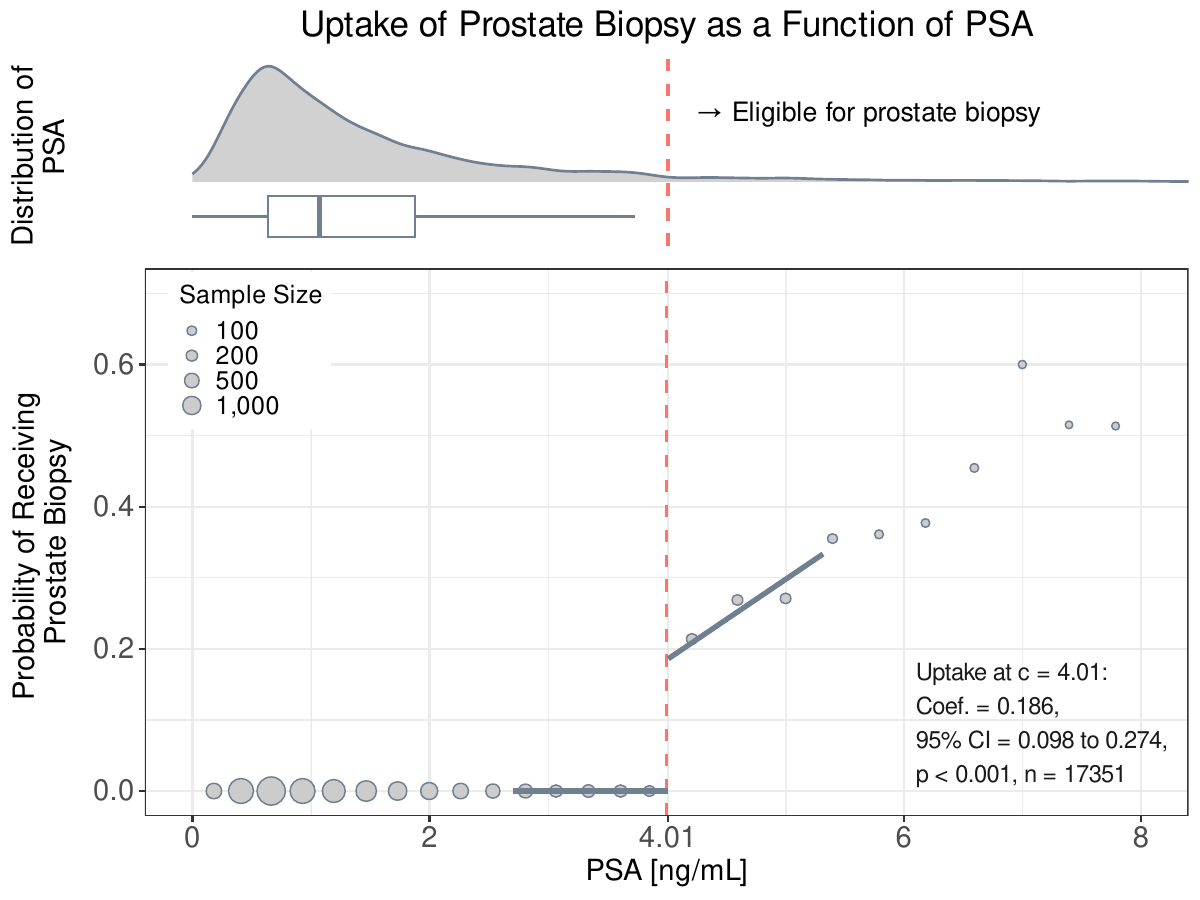}
        \caption[]
        {Uptake of prostate biopsy as a function of the running variable PSA. The density and bar plots (top) shows the distribution and median (with IQR) of PSA. The red dashed line indicates the eligibility threshold for prostate biopsy and the solid lines the linear fits on each side of the threshold (\texttt{rdrobust}).}
        \label{fig:plco_first_stage}
\end{figure}

Our analysis begins by evaluating treatment uptake at the PSA threshold of 4 ng/mL to establish whether a meaningful discontinuity in treatment assignment is present. Figure \ref{fig:plco_first_stage} shows the distribution of PSA values (top) and the uptake in prostate biopsy utilization at the critical threshold (bottom). Because the biopsy threshold of 4 ng/mL lies in the upper tail of the PSA distribution, relatively few patients fall within the vicinity of the threshold. In the first stage, we estimate a 0.177 (95\% CI 0.09 to 0.264, $p < 0.001$) increase in the probability of receiving a prostate biopsy as patients cross the threshold. Although this uptake may appear moderate, a reasonable hypothesis is that many patients may delay their decision to undergo prostate biopsy in this first screening to a later cycle.

\begin{figure}[t]
        \centering
        \includegraphics[width=1.0\textwidth]{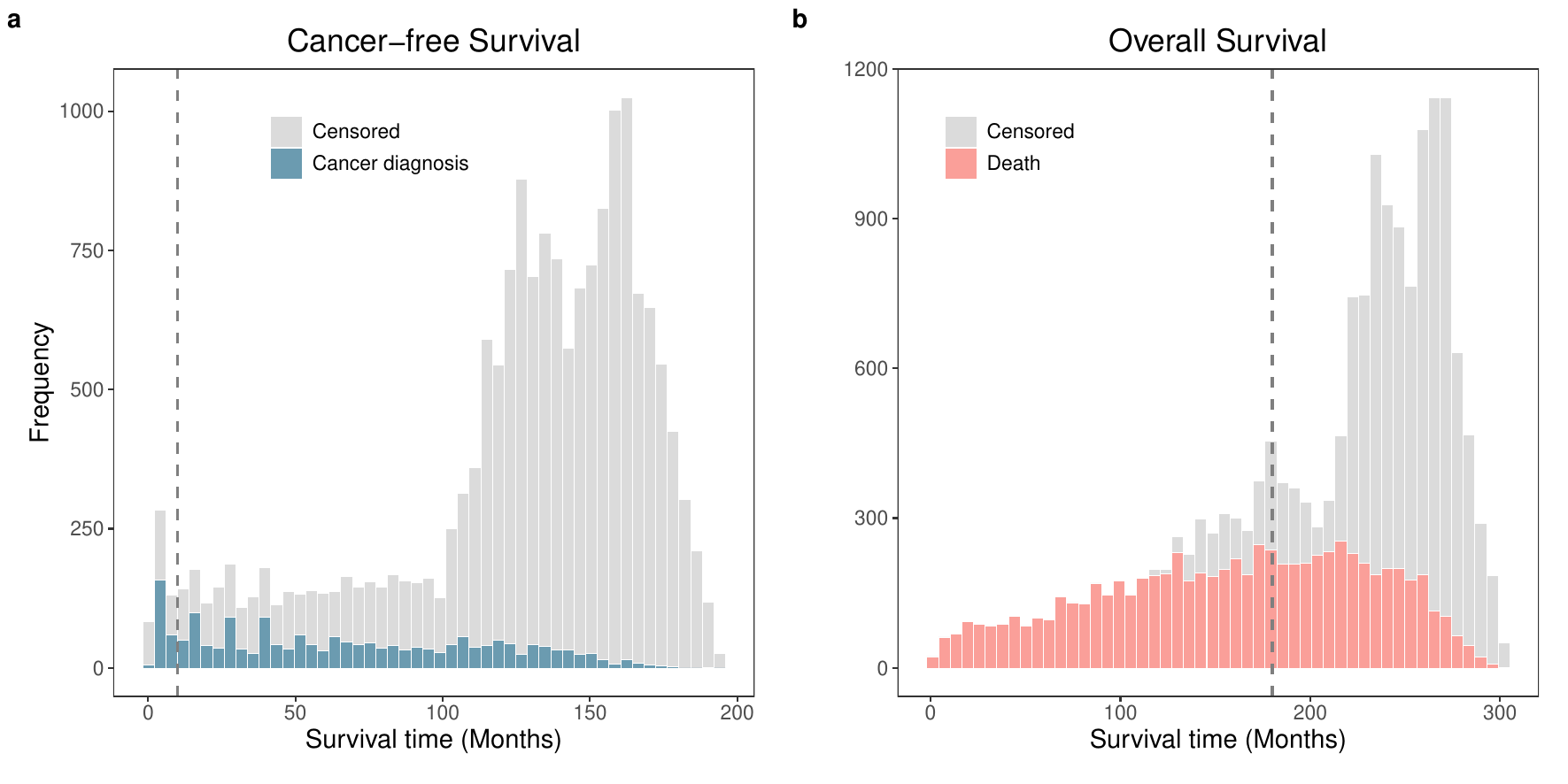}
        \caption[]
        {Event and censoring distributions for prostate cancer-free and overall survival. The gray dashed lines mark the horizons used in our estimation.}
        \label{fig:plco_censoring_event_plot}
\end{figure}

Since screening is repeated annually in the PLCO trial in the first 6 years post-enrollment, we hypothesized that the number of cancer diagnoses might spike several times during the study period. Such periodic upticks with each screening round are plausible because histologic confirmation through biopsy establishes the diagnosis of prostate cancer, thereby resulting in higher cancer detection rates and shortening of the cancer-free interval. Figure \ref{fig:plco_censoring_event_plot} shows that cancer diagnoses spike following each screening cycle and then stabilize over the longer term.

\begin{figure}[p]
        \centering
        \includegraphics[width=1.0\textwidth]{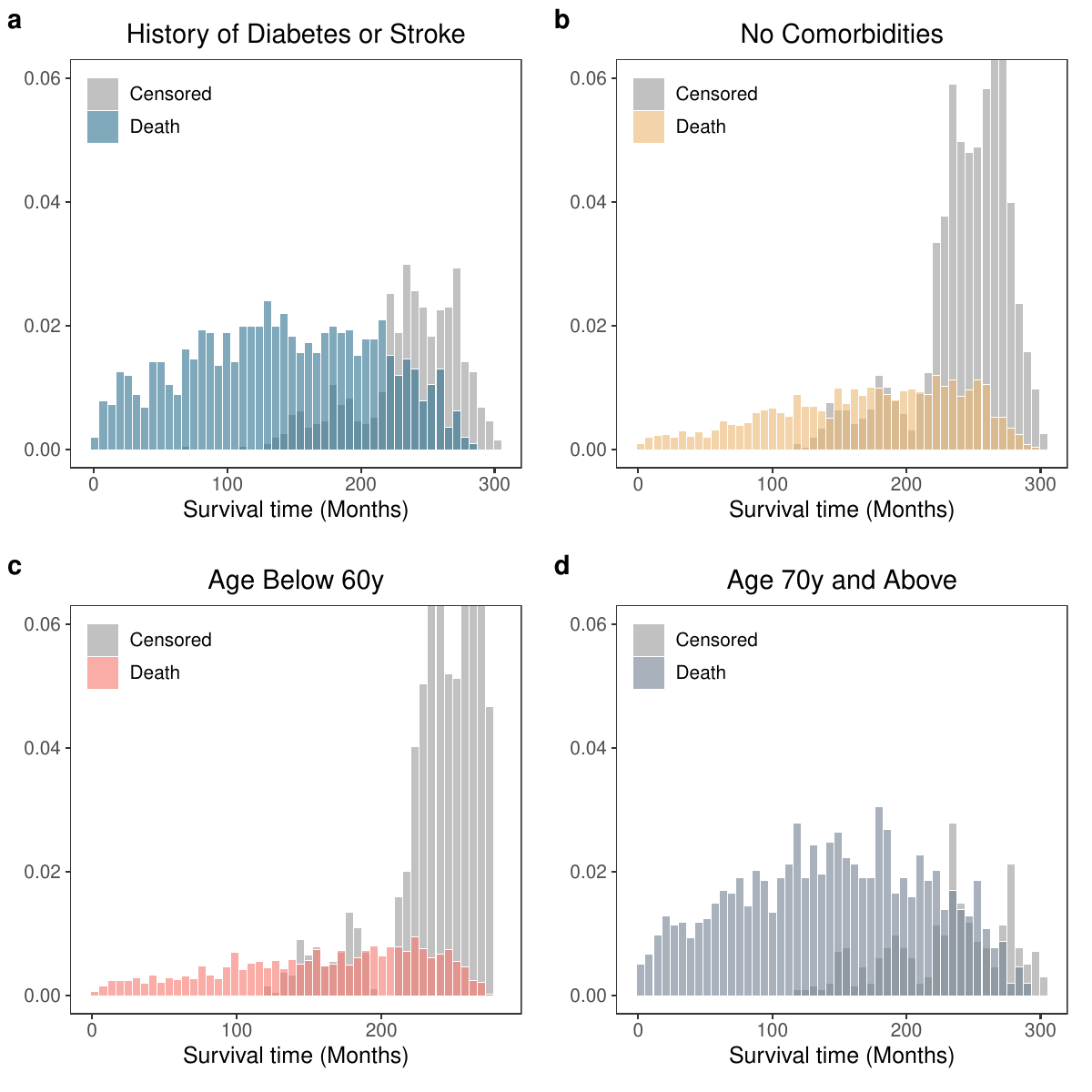}
        \caption[]
        {Distribution of events and right-censoring for overall survival stratified by subgroups. The density plots show the distribution of events and censoring for overall survival (OS).}
        \label{fig:plco_combined_subgroup_panel}
\end{figure}

In contrast, mortality in Figure \ref{fig:plco_censoring_event_plot} b increases as expected over time as participants age and the risk of cancer-related death grows. Early mortality is more common among patients with a history of diabetes or stroke than among those without documented comorbidities at enrollment (Figure \ref{fig:plco_combined_subgroup_panel}a-b). Similarly, event rates are much higher in patients at the age of 70 and above compared to younger cohorts below 60 years (Figure \ref{fig:plco_combined_subgroup_panel}c-d). At the same time, the longer follow-up is associated with higher rates of participant dropout and right-censoring. To formally test for covariate-specific differences in the survival and censoring distributions, we conduct log-rank tests for both the event and censoring outcomes. There are significant differences in event distributions for patients with both stroke and diabetes compared with those without ($p < 0.001$), with at least one comorbidity compared with none ($p < 0.001$), aged 70 or older compared with younger than 70 ($p < 0.001$), and younger than 60 compared with aged 60 or older ($p < 0.001$). These results are consistent for censoring distributions, except for patients with both stroke and diabetes versus those without ($p = 0.135$).
Together, these covariate-dependent survival differences support an approach that is able to incorporate covariates in modeling both censoring and event distributions. Moreover, they are suggestive of distributional patterns in which traditional parametric assumptions such as proportional hazards might not necessarily hold \citep{grambsch1994proportional}. To test this hypothesis, we assessed whether the Cox proportional hazards assumption was met. When investigating individuals around the threshold, the test result revealed that this assumption was violated globally ($\chi^2$-test: $p < 0.001$), as well as at the covariate-specific level, including for age ($p < 0.001$), weight ($p=0.005$), history of coronary artery disease (CAD, $p=0.002$), and years of smoking ($p=0.003$).

\begin{figure}[p]
        \centering
        \includegraphics[width=1.0\textwidth]{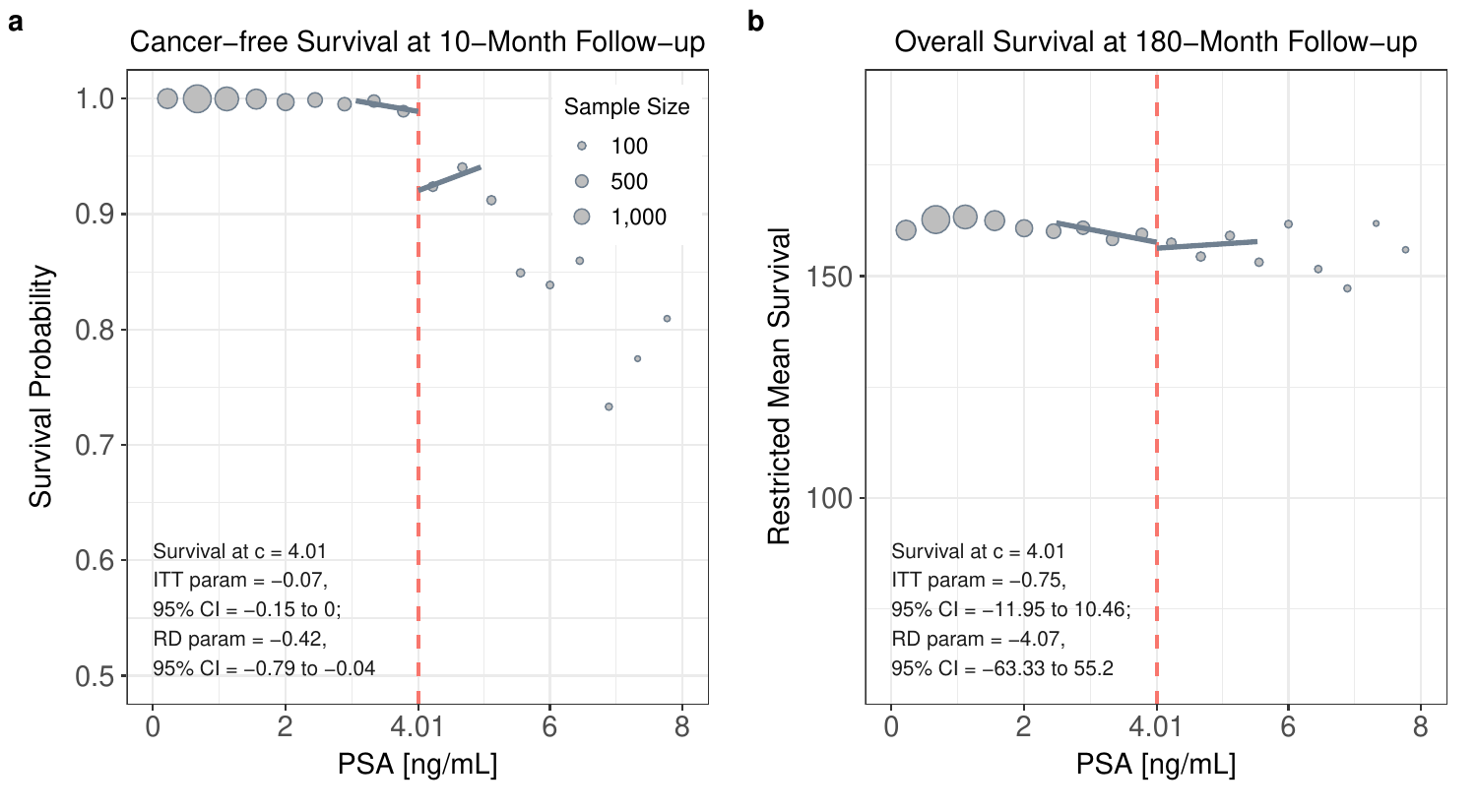}
        \caption[]
        {Effect of prostate biopsy on survival endpoints. The dashed line indicates the eligibility threshold for prostate biopsy. The survival probabilities are estimated using doubly robust scores \eqref{eq:drscores}. The solid lines represent the bias-corrected linear model (\texttt{rdrobust}) fit on each side of the threshold.}
        \label{fig:second_stage_plot_plco}
\end{figure}

A natural first positive control outcome for our RDD is to assess whether receipt of a prostate biopsy at the threshold leads to a shortened prostate cancer-free survival during the first screening cycle. Accordingly, we selected 10 months as our follow-up horizon, i.e. before the second screening takes place. Figure \ref{fig:second_stage_plot_plco} shows the difference in prostate cancer-free survival at 10 months, expressed as a probability. There is a reduction in cancer-free survival, with an intention-to-treat (ITT) effect of -0.07 (95\% CI: -0.15 to 0) and of -0.42 (95\% CI: -0.79 to -0.04) for the RD parameter. This is expected as the biopsy enables the detection of cancers before they become clinically symptomatic.

\begin{figure}[p]
        \centering
        \includegraphics[width=1.0\textwidth]{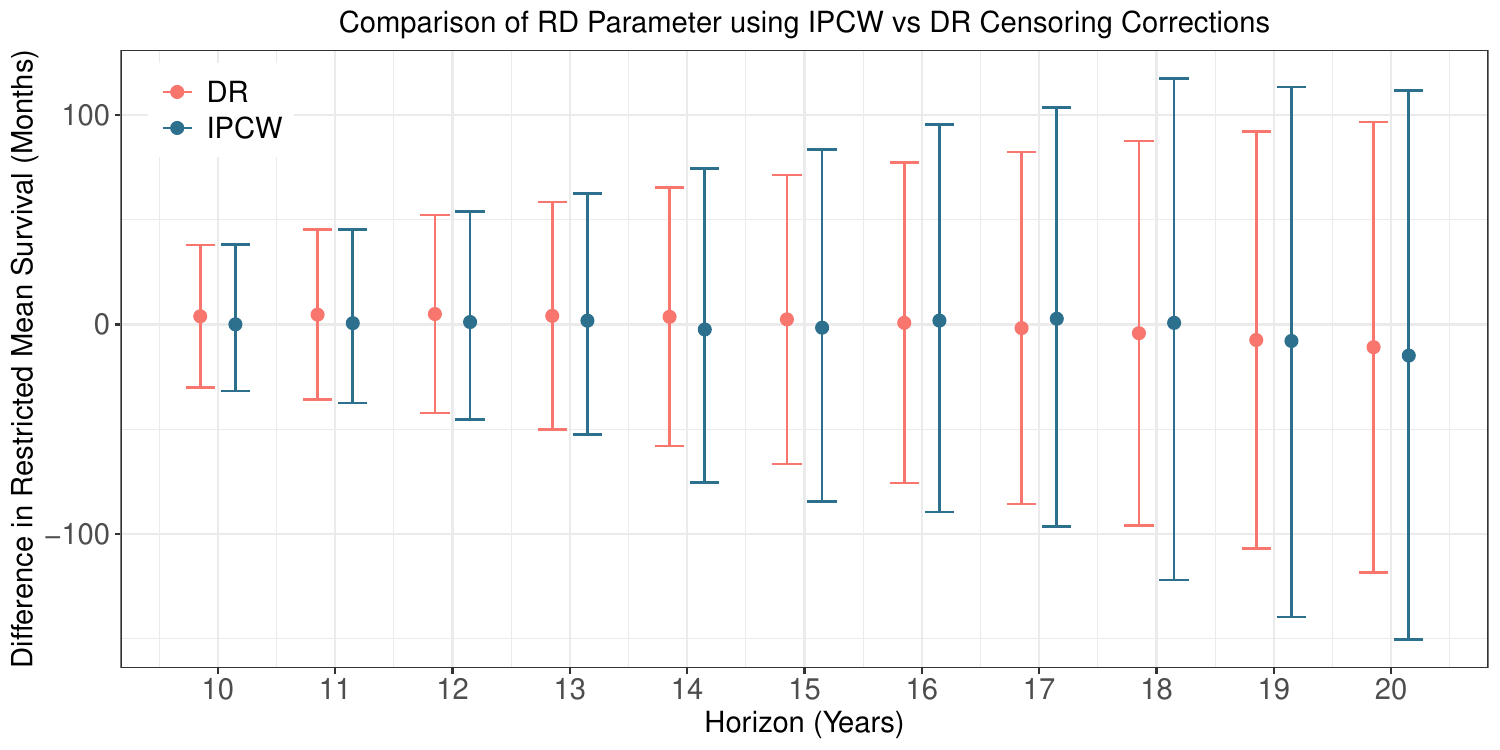}
        \caption[]
        {Effect of prostate biopsy on overall survival at varying time horizons. Point estimates with 95\% confidence intervals for the RD parameter from \texttt{rdrobust} are shown at various time horizons. For each time point, the difference in restricted mean survival time (RMST) when estimated using IPCW and DR scores are presented side-by-side.}
        \label{fig:ipcw_dr_comparison_OS}
\end{figure}

Figure \ref{fig:second_stage_plot_plco}b shows the difference in overall survival at the threshold in restricted mean survival time at 15-year follow-up, with an ITT effect of -0.75 (95\% CI: -11.95 to -10.46) months and RD parameter of -4.07 (95\% CI: -66.33 to 55.22) months, suggesting that there is no significant survival benefit from prostate biopsy on overall survival in PLCO. In a next step, we hypothesized that as censoring rates increase, the DR approach will yield higher precision as it does not discard censored units during the estimation process. Figure \ref{fig:ipcw_dr_comparison_OS} compares the RD Parameter when using an IPCW- versus DR-based approach to address censoring as we increased the horizon (and thus censoring). With increasing time horizons and correspondingly higher censoring rates, both approaches yield comparable point estimates, but the DR-based approach generates tighter confidence intervals.

\begin{figure}[t]
        \centering
        \includegraphics[width=1.0\textwidth]{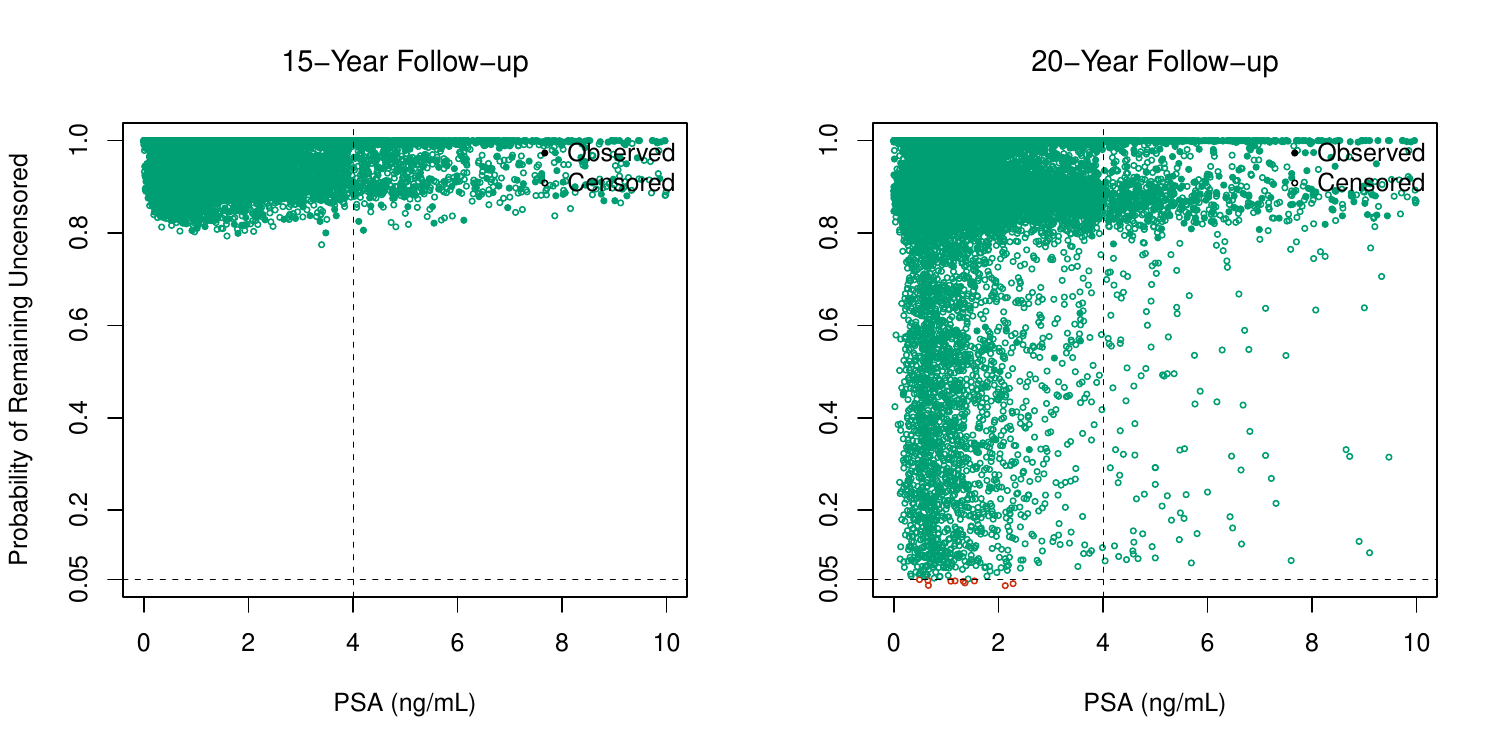}
        \caption[]
        {Relationship between the horizon $h$ and units' probabilities of remaining uncensored when analyzing overall survival. At the horizontal dashed line, the probability of remaining uncensored becomes critically low. Units below this line (red color) and near the threshold (vertical dashed line) may cause instabilities in the DR scores and thus RD estimates.}
        \label{fig:censoring.diagnostics}
\end{figure}

Finally, to test for positivity issues (and thus stability of the RD estimation), we plotted the probability of remaining uncensored against the running variable (Figure \ref{fig:censoring.diagnostics}). At 15 years, most units have probabilities of remaining uncensored $\geq0.8$, indicating no positivity violations. Extending the horizon to 20 years lowers them, but the share of those $\leq0.05$ near the 4 ng/mL threshold remains negligible, confirming that our results pass this diagnostic step.

\section{Discussion}
In this study, we presented a nonparametric, doubly robust approach that addresses right-censoring in RDDs with time-to-event data. We demonstrated the utility of this approach in a large, heterogeneous cohort of the PLCO Cancer Screening Trial for prostate cancer. In this setting, censoring and event distributions were complex and showed variation across demographic groups and participant-level descriptors, supporting the use of a nonparametric approach capable of incorporating covariate information. The application further illustrated how a quasi-experimental design can complement evidence from an RCT, helping to recover valid inference when randomization has been compromised due to contamination of the control group.
The nonparametric, doubly robust properties of our approach are particularly relevant in real-world applications. In healthcare, for example, electronic health records often exhibit fluctuations due to irregular entries and updates, shifts in the distribution of covariates and/or outcomes, and external shocks. Such variation frequently complicates censoring mechanisms, leads to violations of parametric model assumptions, and leaves causal estimates susceptible to bias. In observational settings, where RDDs are most appealing, this is further complicated by censoring rates that are typically much higher than in randomized controlled trials. IPCW-based estimators may then experience notable efficiency losses. One benefit of our strategy is its modular design, allowing users to employ an RD estimator of their choice for each estimand. As new methods for RD estimation emerge, this modularity helps prevent our approach from becoming outdated and supports future methodological extensions.
Future work may explore integrating our approach with other estimator classes, including hazard ratio estimators, and extending it to advanced competing risks settings.
The results from the PLCO trial suggest that prostate biopsy may not confer a statistically significant overall survival benefit at 15 years of follow-up in the PLCO trial cohort of patients who undergo PSA screening for the first time. While this result does not offer a new finding, the context of the PLCO trial as a high-stakes application emphasizes the importance of nonparametric doubly robust approaches to increase confidence in causal estimates. This is particularly relevant when conclusions drawn from causal inference designs, such as regression discontinuity, are intended to inform clinical practice and other areas with potentially life-altering decisions.

\section*{Acknowledgments}
The authors thank the National Cancer Institute for access to NCI's data collected by the Prostate, Lung, Colorectal and Ovarian (PLCO) Cancer Screening Trial (CDAS Project Number: PLCO-1625).  M.Sc. was supported by a Stanford TRAM pilot grant, the Stanford Data Science Scholarship and the MAC3 Computation in Precision Health Fellowship. We would like to thank Stanford University and Stanford Research Computing for providing computational resources and support that contributed to these research results. This work was partially supported by a Stanford Cancer Institute's SCI Cancer Innovation award. R.T. was supported by NIH grant R01 GM134483 and NSF grant 19DMS1208164. S.W. was supported by NSF SES-2242876.

We are grateful to James Brooks for the valuable guidance on the analysis of the PLCO trial. 

\bibliographystyle{plainnat}
\bibliography{bibliography}

\clearpage
\newpage
\appendix

\end{document}